\documentclass[runningheads]{llncs}
\usepackage{graphicx}
\usepackage[dvipsnames]{xcolor}
\usepackage{hyperref}

\usepackage{amsmath}
\usepackage{amsfonts} 
\usepackage[range-units=single]{siunitx}
\usepackage{bm}
\usepackage{multirow}

\begin{document}
\title{Fast Auto-Differentiable Digitally Reconstructed Radiographs for Solving Inverse Problems in Intraoperative Imaging}
\titlerunning{Fast Auto-Differentiable Digitally Reconstructed Radiographs}
\author{Vivek Gopalakrishnan\inst{1,2} \and
Polina Golland\inst{1,2}}
\authorrunning{V. Gopalakrishnan and P. Golland.}
\institute{Harvard-MIT Health Sciences and Technology, Massachusetts Institute of Technology, Cambridge, MA, USA\\
\and
Computer Science and Artificial Intelligence Laboratory, Massachusetts Institute of Technology, Cambridge, MA, USA\\
\email{\{vivekg,polina\}@csail.mit.edu}}

\maketitle
\begin{abstract}
The use of digitally reconstructed radiographs (DRRs) to solve inverse problems such as slice-to-volume registration and 3D reconstruction is well-studied in preoperative settings.
In intraoperative imaging, the utility of DRRs is limited by the challenges in generating them in real-time and supporting optimization procedures that rely on repeated DRR synthesis.
While immense progress has been made in accelerating the generation of DRRs through algorithmic refinements and GPU implementations, DRR-based optimization remains slow because most DRR generators do not offer a straightforward way to obtain gradients with respect to the imaging parameters.
To make DRRs interoperable with gradient-based optimization and deep learning frameworks, we have reformulated Siddon's method, the most popular ray-tracing algorithm used in DRR generation, as a series of vectorized tensor operations.
We implemented this vectorized version of Siddon's method in PyTorch, taking advantage of the library's strong automatic differentiation engine to make this DRR generator fully differentiable with respect to its parameters.
Additionally, using GPU-accelerated tensor computation enables our vectorized implementation to achieve rendering speeds equivalent to state-of-the-art DRR generators implemented in CUDA and {C{}\verb!++!}.
We illustrate the resulting method in the context of slice-to-volume registration.
Moreover, our simulations suggest that the loss landscapes for the slice-to-volume registration problem are convex in the neighborhood of the optimal solution, and gradient-based registration promises a much faster solution than prevailing gradient-free optimization strategies.
The proposed DRR generator enables fast computer vision algorithms to support image guidance in minimally invasive procedures.
Our implementation is publically available at \url{https://github.com/v715/DiffDRR}.
\end{abstract}

\section{Introduction}

Digitally reconstructed radiographs (DRRs) are simulated 2D X-ray images generated from 3D computational tomography (CT) volumes using a variety of ray-tracing techniques.
While DRRs are widely used in preoperative settings (e.g., optimizing dose delivery in radiation oncology), many potentially valuable intraoperative use cases (e.g., real-time multimodal registration for image-guided procedures) are infeasible due to computational bottlenecks in generating DRRs and using them in slice-to-volume registration.
Most open-source CPU-based implementations take about \SI{1}{\sec} to generate a single DRR \cite{sharp2010plastimatch}, which is not fast enough for intraoperative imaging systems with sampling rates of about 7.5 frames per second \cite{Sadamatsu2016TheEO}. 
Numerous GPU-accelerated DRR generators have been proposed with run times on the order of \SI{100}{\m\sec} \cite{de2009accelerated,mori2009development,ruijters2008gpu},
but to the best of our knowledge, no publically available implementation exists.

The second limitation of currently available DRR generators is that it is challenging to efficiently compute derivatives using these renderers because they are implemented in low-level languages such as CUDA and {C{}\verb!++!}.
DRR generators are often used in conjunction with numerical optimization schemes to solve fundamental medical imaging problems (e.g., slice-to-volume registration), and the difficulty in computing derivatives means that gradient-based optimization techniques are often infeasible \cite{van2011evaluation}.
While many end-to-end deep learning approaches can solve X-ray to CT registration problems with high accuracy \cite{esteban2019towards,hou2017predicting}, 
these methods often require large amounts of training data, which can make them impractical for specialized interventional problems.
Instead, many applications use iterative gradient-free methods, such as the Nelder-Mead method \cite{singer2009nelder}, to optimize an image similarity metric with respect to the parameters of the DRR generator \cite{hou2017predicting,van2011evaluation}.
While these methods are effective for optimizing highly nonlinear loss functions, we show that the loss landscapes for slice-to-volume registration in particular is convex in a large region around the optimum, making this problem better suited for gradient-based optimization methods.

We present a fast vectorized renderer that generates DRRs and their derivatives with respect to image geometry parameters automatically. 
We utilize PyTorch as a GPU-accelerated tensor algebra library with robust source-to-source automatic differentiation to implement differentiable DRRs.
That is, using our implementation, DRR generation can be used as a differential operator to train deep learning algorithms for fast reconstruction and registration algorithms.
We analyze the performance of our implementation and the correctness of the automatically obtained derivatives, and demonstrate an experiment where our differentiable DRR generator solves a slice-to-volume registration problem.
Our hope is that this open-source package will be useful for translating computer vision algorithms to real-time implementations for interventional applications.


\section{Methods}
We start by summarizing Siddon's method \cite{siddon1985fast}, commonly used for ray-tracing in DRR synthesis, and its extensions that accelerate rendering speed. 
We then describe our vectorized implementation of Siddon's method, which achieves rendering speeds equivalent to those of existing GPU-accelerated methods while also being fully differentiable.

\begin{figure}[h!!!]
    \centering
    \includegraphics[width=\linewidth]{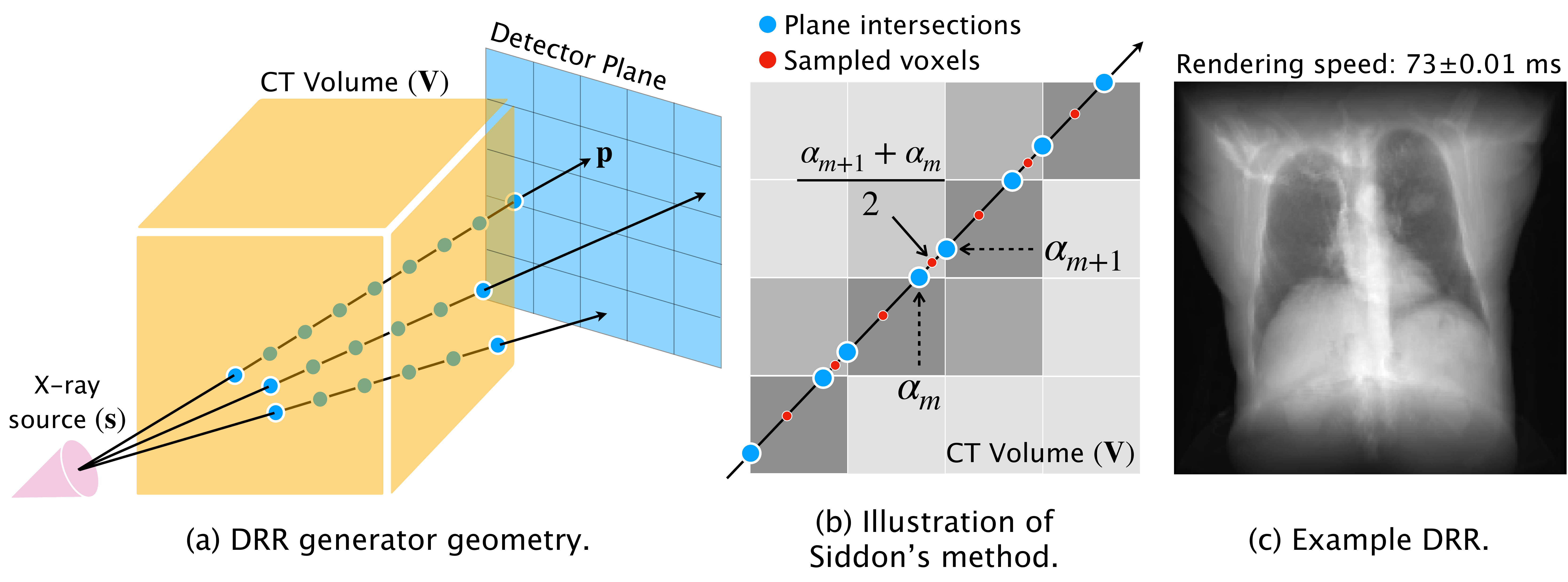}
    \caption{\textbf{DRR synthesis.} (a) We assume an idealized model of a projectional radiography imaging system: X-ray beams are emitted with a fixed initial energy from a point source $\mathbf s \in \mathbb R^3$, beam energy diminishes as the X-rays travel through the CT volume $\mathbf V$, and energy in the attenuated beams is measured when the X-rays hit a point on the detector $\mathbf p \in \mathbb R^3$, producing the DRR. (b) In Siddon's method, the image location value at $\mathbf p$ is a weighted average of the intensities of the voxels through which the ray passes, where the weight is the length of the ray's intersection with the voxel. The values $\alpha_m$ and $\alpha_{m+1}$ parameterize the intersection of the ray with two adjacent planes, and the midpoint $\frac{\alpha_{m+1} + \alpha_m}{2}$ identifies the current voxel through which the ray is passing. (c) Our vectorized Siddon's method generates a $200 \times 200$ DRR in \SI{72.7}{\m\sec} $\pm$ \SI{10}{\micro\sec} on an NVIDIA GeForce RTX 2080 Ti.}
    \label{fig:drr}
\end{figure}

\subsection{DRR Generation}
The process of generating a DRR models the geometry of an idealized projectional radiography system (Fig. \ref{fig:drr}a).
Let $\mathbf s \in \mathbb R^3$ be the X-ray source, and $\mathbf p \in \mathbb R^3$ be a target pixel on the detector plane.
Then $R(\alpha) = \mathbf s + \alpha (\mathbf p - \mathbf s)$ is a ray that originates from $\mathbf s$ ($\alpha=0$), passes through the imaged volume, and hits the detector plane at $\mathbf p$ ($\alpha=1$).
The total energy attenuation experienced by the X-ray by the time it reaches pixel $\mathbf p$ is given by the following line integral:
\begin{equation}
    \label{eqn:line-int}
    E(R) = \|\mathbf p - \mathbf s\|_2 \int_0^1 \mathbf V \left( \mathbf s + \alpha (\mathbf p - \mathbf s) \right) \mathrm d\alpha ,
\end{equation}
where $\mathbf V : \mathbb R^3 \mapsto \mathbb R$ is the imaged volume.
The term $\|\mathbf p - \mathbf s\|_2$ endows the unit-free $\mathrm d \alpha$ with the physical unit of length.
For DRR synthesis, $\mathbf V$ is approximated by a discrete 3D CT volume, and Eq. (\ref{eqn:line-int}) becomes
\begin{equation}
    \label{eqn:radiologic-path-length}
    E(R) = \|\mathbf p - \mathbf s\|_2 \sum_{m=1}^{M-1} (\alpha_{m+1} - \alpha_m) \mathbf V \left[ \mathbf s + \frac{\alpha_{m+1} + \alpha_m}{2} (\mathbf p - \mathbf s) \right] ,
\end{equation}
where $\alpha_m$ parameterizes the locations where ray $R$ intersects one of the orthogonal planes comprising the CT volume, and $M$ is the number of such intersections (Fig. \ref{fig:drr}b).
Note that this model does not account for patterns of reflection and scattering that are present in real X-ray systems.
While these simplifications preclude synthesis of realistic X-rays, the model in Eq. (\ref{eqn:radiologic-path-length}) has been widely and successfully used in slice-to-volume registration \cite{van2011evaluation}.
Additionally, our approach of vectorizing DRR generation might also be interoperable with more sophisticated image synthesis models, an extension we examine further in the Discussion.

\subsection{Siddon's Method and Its GPU Extensions}
Siddon's method \cite{siddon1985fast} provides a parametric method to identify the plane intersections $\{\alpha_m\}_{m=1}^M$.
Let $\Delta X$ be the CT voxel size in the $x$-direction and $b_x$ be the location of the $0$-th plane in this direction.
Then the intersection of ray $R$ with the $i$-th plane in the $x$-direction is given by
\begin{equation}
    \label{eqn:x-intersect}
    \alpha_x(i) = \frac{b_x + i \Delta X - \mathbf s_x}{\mathbf p_x - \mathbf s_x} ,
\end{equation}
with analogous expressions for $\alpha_y(\cdot)$ and $\alpha_z(\cdot)$.
We can use Eq. (\ref{eqn:x-intersect}) to compute the values $\bm \alpha_x$ for all the intersections between $R$ and the planes in the $x$-direction:
\begin{equation*}
    \bm\alpha_x = \{ \alpha_x(i_{\min}), \dots, \alpha_x(i_{\max}) \} ,
\end{equation*}
where $i_{\min}$ and $i_{\max}$ denote the first and last intersections of $R$ with the $x$-direction planes.
Defining $\bm\alpha_y$ and $\bm\alpha_z$ analogously, we construct the array
\begin{equation}
    \label{eqn:alphas}
    \bm\alpha = \mathrm{sort}(\bm\alpha_x, \bm\alpha_y, \bm\alpha_z) ,
\end{equation}
which contains $M$ values of $\alpha$ parameterizing the intersections between $R$ and the orthogonal $x$-, $y$-, and $z$-directional planes. 
We substitute values in the sorted set $\bm\alpha$ into Eq. (\ref{eqn:radiologic-path-length}) to evaluate $E(R)$, which corresponds to the intensity of pixel $\mathbf p$ in the synthesized DRR.

A faster variant determines consecutive intersecting planes iteratively \cite{jacobs1998fast}.
For example, the value of $\alpha$ at the second plane intersected by $R$ is given by $\alpha_2 = \min\{\alpha_x(i_{\min}+1), \alpha_y(j_{\min}+1), \alpha_z(k_{\min}+1)\}$.
The algorithm iteratively finds the next value of $\alpha$ until we reach the edge of the CT volume, making this approach more memory efficient by requiring fewer intermediate values to be stored.
This modified algorithm, known as Siddon-Jacobs' method, is commonly implemented in CUDA and {C{}\verb!++!} to create multi-threaded GPU-accelerated DRR generators that exploit data parallelism by assigning each thread to trace an independent ray intersecting the detector plane \cite{de2009accelerated,mori2009development,ruijters2008gpu}.

\subsection{Vectorizing Siddon's Method}

While Siddon-Jacobs' method is more memory efficient, the iterative loop it relies on is not amenable to implementations in vectorized tensor algebra libraries.
Thus we vectorize the original Siddon's method as follows.
Let $\mathbf P \in \mathbb R^{H \times W \times 3}$ contain the 3D pixel coordinates of a DRR with dimension $H \times W$.
We compute the $\alpha$ values for intersections with all of the $x$-, $y$-, and $z$-planes for all $\mathbf p \in \mathbf P$ in parallel:
\begin{equation}
    \mathbf A
    = \left(
        \begin{pmatrix} b_x \\ b_y \\ b_z \end{pmatrix}
        + \begin{pmatrix} i \\ j \\ k \end{pmatrix}
        \otimes \begin{pmatrix} \Delta X \\ \Delta Y \\ \Delta Z \end{pmatrix}
        - \mathbf s
    \right)
    \oslash (\mathbf P - \mathbf s) \in \mathbb R^{H \times W \times (n_x + n_y + n_z)} ,
\end{equation}
where $(n_x, n_y, n_z)$ are the dimensions of the CT volume $\mathbf V$, $(i,j,k)$ are the CT voxel indices, $(\Delta X, \Delta Y, \Delta Z)$ are the CT voxel sizes, and $\otimes$ and $\oslash$ are the Hadamard product and division operators, respectively.
Rather than explicitly compute the indices $(i_{\min},i_{\max})$, $(j_{\min},j_{\max})$, and $(k_{\min},k_{\max})$ for each ray, as is done in Siddon's original method, we instead compute the minimum and maximum values of $\bm\alpha$, corresponding to when each ray enters and exits the volume:
\begin{align*}
    \bm\alpha_{\min} &= \max\big\{ \min\{\bm\alpha_x(0), \bm\alpha_x(n_x)\}, \min\{\bm\alpha_y(0), \bm\alpha_y(n_y)\}, \min\{\bm\alpha_z(0), \bm\alpha_z(n_z)\} \big\} \\
    \bm\alpha_{\max} &= \min\big\{ \max\{\bm\alpha_x(0), \bm\alpha_x(n_x)\}, \max\{\bm\alpha_y(0), \bm\alpha_y(n_y)\}, \max\{\bm\alpha_z(0), \bm\alpha_z(n_z)\} \big\} ,
\end{align*}
where $\bm\alpha_{\min}, \bm\alpha_{\max} \in \mathbb R^{H \times W}$.
We filter $\mathbf A$ to include only  values in the range $[\bm\alpha_{\min}, \bm\alpha_{\max}]$ and sort each row $\mathbf A(h,w,\cdot)$ for $h\in\{1,\dots,H\}, w\in\{1,\dots,W\}$.
Finally, we evaluate Eq. (\ref{eqn:radiologic-path-length}) with this sorted tensor to compute the intensity for each pixel in the DRR, completing a chain of vectorized tensor operations. 

Because we reformulated the original Siddon's method as a series of tensor operations, our vectorized version benefits from the mature GPU compilers and memory allocators developed for optimizing large-scale deep learning models.
For empirical evaluation of our method, we also implemented a partially-vectorized version of Siddon-Jacobs' method in which the $\alpha$ updates are still computed iteratively (i.e., with a loop), but the updates are applied in a vectorized form to every target pixel in the detector plane.

\subsection{Differentiating DRRs with Respect to Imaging Parameters}
We specify the 3D position of the X-ray source and detector plane relative to the CT volume with the following seven geometric parameters: radius $\rho$ that acts as a scaling factor; three rotational degrees of freedom (DoF) $(\theta, \varphi, \gamma)$; and three translational DoF $(b_x, b_y, b_z)$.
Using spherical coordinates, we express the position of the X-ray source as $\mathbf s = (\rho, \theta, \varphi)$, where $\rho$ is half of the source-to-detector distance, and $\theta$ and $\varphi$ are the azimuthal and polar angles, respectively.
We assume the detector plane is tangent to this implied sphere at the point opposite $\mathbf s$.
The orientation of this plane is determined by a rotation about the $x$-axis by the angle $\gamma$.
We add the translation $(b_x, b_y, b_z)$ to the coordinates of the X-ray source and detector plane to create a reference frame wherein the patient is not perfectly centered relative to the X-ray scanner.

Since every step of our pipeline, from the generation of the pixels on the detector plane to the computation of the pixel intensities, is performed in PyTorch's tensor framework, the resulting DRRs are differentiable with respect to the parameters described above.
That is, the gradient $\nabla_{\bm\eta} \mathcal L(I(\bm\eta))$ of the loss function $\mathcal L(\cdot)$ evaluated for the DRR $I(\bm\eta)$ with respect to the parameters $\bm\eta=(\rho, \theta, \varphi, \gamma, b_x, b_y, b_z)$ is obtained automatically for any differentiable $\mathcal L(\cdot)$.

\section{Experiments}

We evaluate the proposed efficient implementation of the algorithm on a reference chest CT scan from Slicer3D \cite{pieper20043d}.
We compare the performance of the method to two baseline approaches and illustrate its application for gradient-based slice-to-volume registration.

\subsection{Performance Analysis}
We compare our vectorized GPU version of Siddon's method (VGS) to two baseline approaches: a widely used CPU implementation in the Plastimatch package (CP) \cite{sharp2010plastimatch}, and our vectorized GPU implementation of Siddon-Jacobs' method (VGSJ) which we described in Section 2.3.
Note that the CUDA-accelerated DRR generator in Plastimatch is not working at the time of publication.
GPU benchmarks are run on an NVIDIA GeForce RTX 2080 Ti, and CPU benchmarks are run on an 18-core Linux computer with Intel(R) Xeon(R) CPU E5-2697 v4 @ 2.30GHz processors. 

\subsubsection{Results.}
Table~\ref{tab:timing} summarizes statistics for the run time and accuracy of our method and the two baseline approaches, as well as intensity differences between our method and Plastimatch, and gradient differences between our method and FFD.
Our implementation is much faster than Plastimatch, which is to be expected as Plastimatch is executed on the CPU. Numerically, the two implementations are very similar with an average root-mean-square error (RMSE) of (8.3$\pm$1.9) $\times 10^{-4}$ where the images are normalized to the range of $[0,1]$.
For DRRs smaller than $H=500$, our method is faster than the vectorized version of Siddon-Jacobs' despite our method's high memory requirements. However, at $H=500$, they have roughly equivalent run times. For smaller image sizes, our DRR generator achieves equivalent run times to previously reported GPU-accelerated implementations \cite{de2009accelerated,mori2009development,ruijters2008gpu}.
The gradients obtained via PyTorch auto-differentiation for our method are within 0.05$\pm$0.01 of those computed via forward finite differences with a step size of $10^{-6}$ and are an order of magnitude more efficient to generate (35.1 ms ± 73.3 µs \textit{vs} 400.5 ms ± 821.4 µs).

\begin{table}[]
\caption{\textbf{Benchmark results.} The dimension of the DRRs is $H \times W$. Each metric is averaged over $20$ runs. (VGS = Vectorized GPU Siddon's method, VGSJ = Vectorized GPU Siddon-Jacobs' method, and CP = CPU Plastimatch.)}
\label{tab:timing}
\begin{tabular}{c||ccc||c||c}
 & \multicolumn{3}{c||}{Timing (ms)} & RMSE & Autograd \textit{vs} FFD \\ \hline
$H=W$ & \multicolumn{1}{c|}{VGS} & \multicolumn{1}{c|}{VGSJ} & CP & VGS \textit{vs} CP & VGS \\ \hline
100 & \multicolumn{1}{c|}{\textbf{17.6 $\pm$ 0.05}} & \multicolumn{1}{c|}{380 $\pm$ 20.9} & 1028 $\pm$ 132 & (6.9$\pm$2.2)$\times10^{-4}$ & 0.03 ± 0.02 \\
200 & \multicolumn{1}{c|}{\textbf{72.7 $\pm$ 0.01}} & \multicolumn{1}{c|}{424 $\pm$ 4.2} & 1784 $\pm$ 488 & (8.7$\pm$2.7)$\times10^{-4}$ & 0.06 ± 0.01 \\
300 & \multicolumn{1}{c|}{\textbf{165 $\pm$ 0.13}} & \multicolumn{1}{c|}{432 $\pm$ 19.2} & 2941 $\pm$ 821 & (6.4$\pm$1.2)$\times10^{-4}$ & 0.08 ± 0.02 \\
400 & \multicolumn{1}{c|}{\textbf{296 $\pm$ 0.06}} & \multicolumn{1}{c|}{425 $\pm$ 2.8} & 6472 $\pm$ 643 & (9.0$\pm$1.9)$\times10^{-4}$ & 0.03 ± 0.005 \\
500 & \multicolumn{1}{c|}{453 $\pm$ 41.2} & \multicolumn{1}{c|}{\textbf{425 $\pm$ 4.6}} & 8472 $\pm$ 478 & (11.7$\pm$4.3)$\times10^{-4}$ & 0.07 ± 0.006
\end{tabular}
\end{table}

\subsection{DRR-based Gradient Descent for Slice-to-Volume Registration}

We use our auto-differentiable DRR generator to implement slice-to-volume registration with synthetic DRRs.
Specifically, we generate a fixed DRR from a set of ground truth parameters $\bm\eta^* = (\theta, \varphi, \gamma, b_x, b_y, b_z)$, and generate a second moving DRR from a set of random initial parameters $\bm\eta_0$.
We use basic gradient descent to minimize the negative Zero-Normalized Cross-Correlation (ZNCC) between the fixed DRR and the moving DRR.

\subsubsection{Image Similarity Metrics are Locally Convex.}
\begin{figure}[h!!!]
    \centering
    \includegraphics[width=\linewidth]{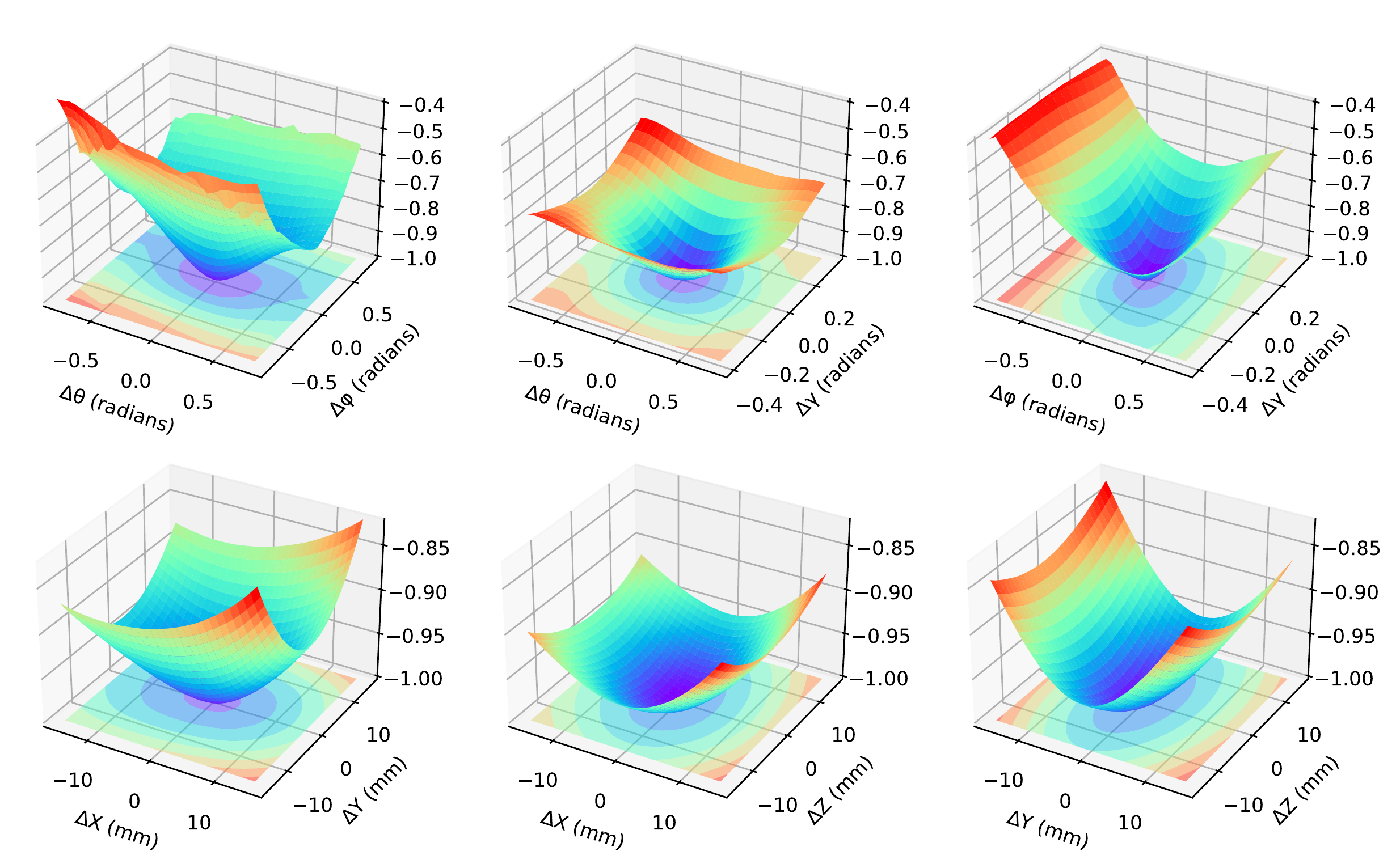}
    \caption{\textbf{Negative ZNCC is convex around the optimal DRR parameters.} Rotational and positional displacements were sampled uniformly from ranges of $90^\circ$ for $\theta$ and $\varphi$, $45^\circ$ for $\gamma$, and \SI{30}{\mm} for $b_x$, $b_y$, and $b_z$.}
    \label{fig:loss-landscape}
\end{figure}

First, we conduct a simulation study to show that the loss landscape generated by negative ZNCC is convex in a neighborhood around $\bm\eta^*$.
We generate moving DRRs by sampling rotational and positional displacements.
We sample all parameters uniformly from ranges of $90^\circ$ for $\theta$ and $\varphi$, $45^\circ$ for $\gamma$, and \SI{30}{\mm} for $b_x$, $b_y$, and $b_z$ around the ground truth parameters $\bm \eta^*$.
Although our generator provides gradients with respect to other model parameters like the source-to-detector distance ($2\rho$) and the DRR's dimensions and pixel spacing ($H, W, \Delta x, \Delta y$), we assume that those parameters are fixed (e.g., provided in the DICOM header), and instead focus our analysis on this 6 DoF registration problem.
We observe that negative ZNCC is locally convex (Fig. \ref{fig:loss-landscape}), suggesting that it would be an apt loss function to optimize with gradient descent.
We observed similar loss landscapes for the $L^2$ norm.

\begin{figure}[h!!!]
    \centering
    \includegraphics[width=0.99\linewidth]{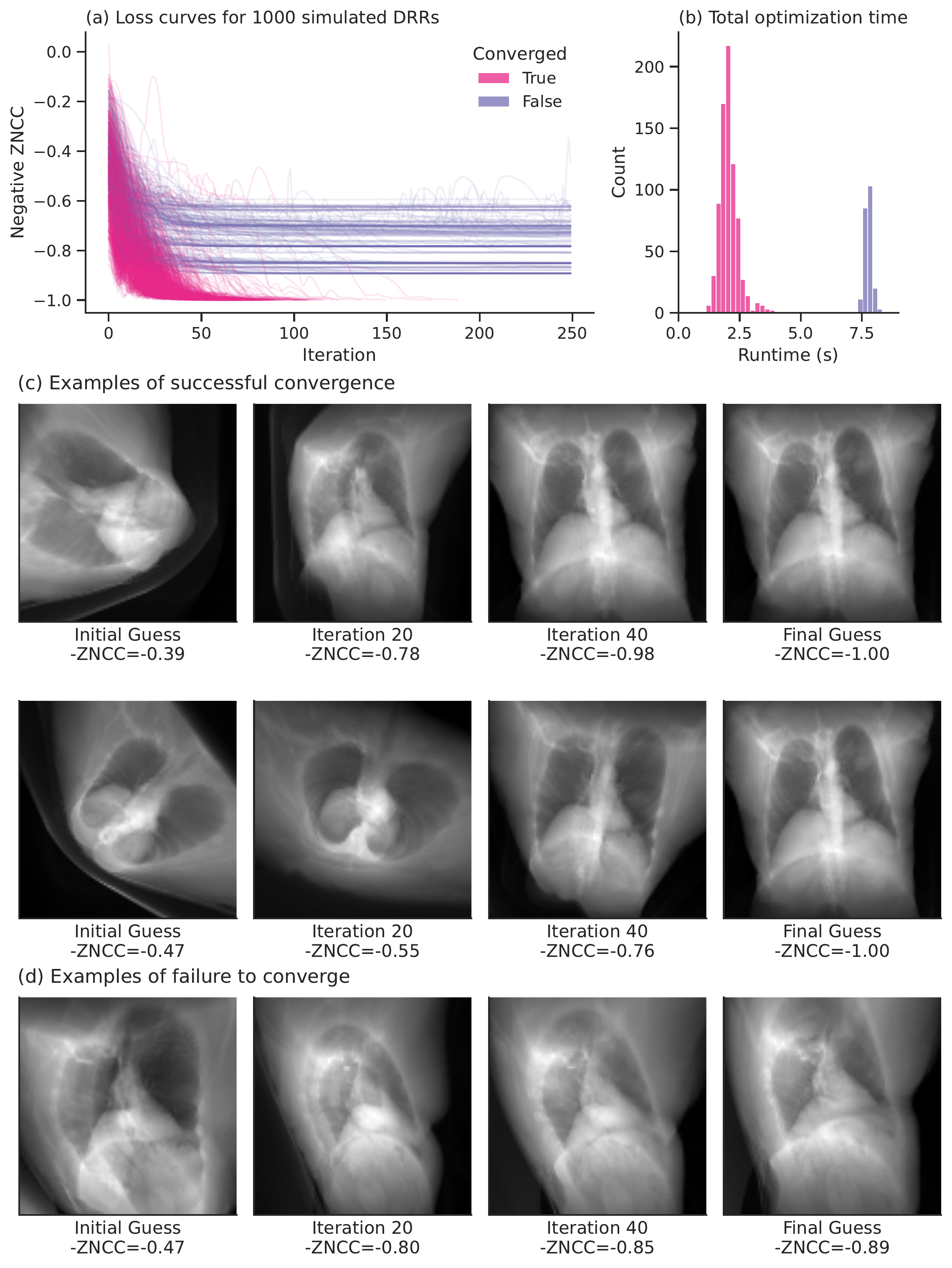}
    \caption{\textbf{Differentiable DRRs can be used to perform slice-to-volume registration.} We generated a moving DRR from randomly initialized parameters and used gradient descent to maximize similarity with a fixed DRR. Convergence was achieved for 745/1000 simulated DRRs in an average 66 iterations (\SI{1.92}{\sec}). Examples of the optimization process are visualized at initial, intermediate (the 20th and 40th iterations), and final steps. DRRs for which convergence fails to occur get stuck in a local minimum with low negative ZNCC.}
    \label{fig:optim}
\end{figure}

\clearpage 
\subsubsection{Differentiable DRR Registration Converges Quickly.}
Given a fixed DRR and a moving DRR, we optimized the parameters of the moving DRR with a basic implementation of gradient descent. We used different update rates for the rotational and translational parameters because they have different units ($\beta_{\theta\varphi\gamma}=5.3\times10^{-2}$ and $\beta_{xyz}=7.5\times10^{1}$), and momentum $\lambda = 0.9$.
Additionally, to investigate the local minima in which our gradient descent algorithm could get stuck, we expanded the space of possible initializations beyond a convex neighborhood to $120^\circ$ for $\theta$, $\varphi$, and $\gamma$, and \SI{60}{\mm} for $b_x$, $b_y$, and $b_z$.
The size of the CT volume is \SI{360}{\mm} $\times$ \SI{360}{\mm} $\times$ \SI{332.5}{\mm}.

For each randomly initialized DRR, we ran 250 iterations of gradient descent and determined that the moving DRR had converged to the fixed DRR if the negative ZNCC between the two images was less than $-0.999$.
If this did not occur within 250 iterations, we treated the run as having failed to converge.
Of the 1,000 randomly initialized DRRs we generated, 745 converged and 255 failed to converge (Fig. \ref{fig:optim}a).
The initializations that converged solved the 6 DoF slice-to-volume registration problem $65.48 \pm 14.27$ iterations (\SI{1.92}{} $\pm$ \SI{0.43}{\sec}).

We visualize multiple optimization steps for our gradient descent registration algorithm (Fig. \ref{fig:optim}c). From the examples that converged, our model successfully recovers the true pose parameters from challenging initializations, reaching a more reasonable estimate by the intermediate 20th iteration (Fig. \ref{fig:optim}b).
In one example of an initialization that failed to converge, we see that the model gets stuck in a local minimum (Fig. \ref{fig:optim}c).
In the final iteration in this example, the estimated DRR is orthogonal to the sagittal plane.
The geometry of this scene looks similar to the coronal DRR (ground truth), giving the illusion of two lungs.
We emphasize that our goal is not to propose a novel registration algorithm, but to provide an efficient DRR synthesis procedure that can support numerous downstream optimization applications, including registration.

\section{Discussion}
We present a fast auto-differentiable DRR generator that can be used to solve inverse problems in intraoperative imaging. 
By reformulating Siddon's method for ray-tracing through a CT volume as series of vectorized tensor operations, we obtain gradients of image loss functions with respect to DRR generating parameters while achieving rendering speeds equivalent to multi-threaded GPU-accelerated generators written in low-level languages such as CUDA and {C{}\verb!++!}.
This approach promises to enable fast solutions to DRR-based optimization problems with gradient methods, a strategy which was previously infeasible due to the inefficiency of generating gradients with finite differences.
We demonstrate the effectiveness of this approach by solving a 6 DoF slice-to-volume registration problem using a locally convex image loss function.

Our future research on auto-differentiable DRRs will investigate their use in a deep learning framework to pre-train for specific intraoperative imaging tasks (e.g., spatiotemporal registration in interventional cardiology).
Such pre-training could yield even faster solutions to inverse problems by finding better initializations or powering end-to-end models. One issue that might limit the effectiveness of such pre-training is that DRRs generated using Siddon's method are not realistic because they do not model any form of scattering. We will investigate fusing our method with DeepDRR \cite{unberath2018deepdrr}, a deep learning framework that estimates scattering effects using Monte Carlo simulations, to produce DRRs that are simultaneously realistic, fast, and differentiable.

\bigskip
\noindent
\textbf{Acknowledgements.}
The authors thank Ruby Liu and Allie Forman for helpful feedback and support.
This work was supported, in part, by NIH NIBIB 5T32EB1680, NIH NIBIB NAC P41EB015902, and NIH NINDS U19NS115388.

\bibliographystyle{splncs04}
\bibliography{references}
\end{document}